\documentclass[letterpaper]{article} % DO NOT CHANGE THIS
\usepackage{aaai2026}  % DO NOT CHANGE THIS
\usepackage{times}  % DO NOT CHANGE THIS
\usepackage{helvet}  % DO NOT CHANGE THIS
\usepackage{courier}  % DO NOT CHANGE THIS
\usepackage[hyphens]{url}  % DO NOT CHANGE THIS
\usepackage{graphicx} % DO NOT CHANGE THIS
\usepackage{natbib}  % DO NOT CHANGE THIS AND DO NOT ADD ANY OPTIONS TO IT
\usepackage{caption} % DO NOT CHANGE THIS AND DO NOT ADD ANY OPTIONS TO IT
\usepackage{algorithm}
\usepackage{algorithmic}

\usepackage{newfloat}
\usepackage{listings}
\DeclareCaptionStyle{ruled}{labelfont=normalfont,labelsep=colon,strut=off} % DO NOT CHANGE THIS
\floatstyle{ruled}
\newfloat{listing}{tb}{lst}{}
\floatname{listing}{Listing}
\usepackage{amsmath}
\usepackage{amssymb}
\usepackage{mathtools}
\usepackage{tabularray}
\UseTblrLibrary{diagbox}
\usepackage{rotating}
\usepackage{hhline}
\usepackage{adjustbox}
\usepackage{makecell}
\usepackage{multirow}
\usepackage{xcolor}
\usepackage{tcolorbox}

\usepackage{times}
\usepackage{helvet}
\usepackage{courier}

\DeclareMathOperator*{\argmin}{arg\,min}

\newcommand{\Mmat}[0]{{{\boldsymbol M}}}

\newcommand{\Wmat}[0]{{{\boldsymbol W}}}
\newcommand{\Xmat}{{\boldsymbol X}}
\newcommand{\Ymat}[0]{{{\boldsymbol Y}}}
\newcommand{\Zmat}{{\boldsymbol Z}}

\title{3One2: One-step Regression Plus One-step Diffusion for One-hot Modulation in Dual-path Video Snapshot Compressive Imaging}

\author{
    Ge Wang\textsuperscript{\rm 1,2},
    Xing Liu\textsuperscript{\rm 3},
    Xin Yuan\textsuperscript{\rm 2,3,}\thanks{Corresponding author}\\
}
\affiliations{
    \textsuperscript{\rm 1}Zhejiang University, Hangzhou, Zhejiang 310058, China\\
    \textsuperscript{\rm 2}Westlake University, Hangzhou, Zhejiang 310030, China\\
    \textsuperscript{\rm 3}Westlake Institute for Optoelectronics, Hangzhou, Zhejiang 311421, China\\
    
    \{wangge,xyuan\}@westlake.edu.cn
    
}

\usepackage{bibentry}
\begin{document}

\maketitle

\begin{abstract}
% Video Snapshot Compressive Imaging (SCI) captures dynamic scene sequences through a two-dimensional (2D) snapshot, fundamentally relying on optically modulated hardware compression and the corresponding software reconstruction. Mature optical design and advanced deep learning algorithms enable the practical application of video SCI. However, several challenges persist: 1) Temporal information capture entails spatial resolution sacrifice during modulated compression. 2) Multiplexing in compressive acquisition results in reduced dynamic range. To address the second challenge, we employ one-hot modulation to derive full-dynamic-range measurements and transform the original regression-based reconstruction into diffusion-based inpainting by introducing a stochastic differential equation (SDE), whose forward process aligns with the hardware compression process. We then utilize one-step regression combined with one-step diffusion to achieve efficient reconstruction. Furthermore, to tackle the first challenge, we introduce a dual optical path that enhances the inpainted video using complementary information from another path. To the best of our knowledge, this approach is the first to integrate the diffusion method into video SCI reconstruction. Experimental results on synthetic and real data validate the effectiveness of the proposed method under one-hot setting; equipped with pixel enhancement, our method achieves state-of-the-art results across various metrics.

Video snapshot compressive imaging (SCI) captures dynamic scene sequences through a two-dimensional (2D) snapshot, fundamentally relying on optical modulation for hardware compression and the corresponding software reconstruction. While mainstream video SCI using random binary modulation has demonstrated success, it inevitably results in temporal aliasing during compression. One-hot modulation, activating only one sub-frame per pixel, provides a promising solution for achieving perfect temporal decoupling, thereby alleviating issues associated with aliasing. However, no algorithms currently exist to fully exploit this potential. To bridge this gap, we propose an algorithm specifically designed for one-hot masks. First, leveraging the decoupling properties of one-hot modulation, we transform the reconstruction task into a generative video inpainting problem and introduce a stochastic differential equation (SDE) of the forward process that aligns with the hardware compression process. Next, we identify limitations of the pure diffusion method for video SCI and propose a novel framework that combines one-step regression initialization with one-step diffusion refinement. Furthermore, to mitigate the spatial degradation caused by one-hot modulation, we implement a dual optical path at the hardware level, utilizing complementary information from another path to enhance the inpainted video. To our knowledge, this is the first work integrating diffusion into video SCI reconstruction. Experiments conducted on synthetic datasets and real scenes demonstrate the effectiveness of our method.
\end{abstract}


\section{Introduction}

Capturing high-speed temporal events is crucial across various scientific and technological fields, including fluid dynamics \cite{adrian2011particle}, biomechanics \cite{lieberman2010foot}, and materials science \cite{parab2019investigation}. Conventional high-speed camera imaging techniques face prohibitive hardware and storage transmission costs. Inspired by compressive sensing \cite{candes2006robust,donoho2006compressed}, video snapshot compressive imaging (SCI) \cite{yuan2021snapshot} employs optical hardware to multiplex a sequence of video frames, each encoded with a distinct modulation pattern (hereafter referred to as a mask), into a single snapshot measurement on a two-dimensional (2D) detector. Additionally, it incorporates hardware-adapted algorithms to reconstruct the original video scene. Within this hardware-encoder plus software-decoder framework, video SCI system achieves orders of magnitude improvements in temporal resolution.
% while deep learning-based reconstruction effectively bridges the gap between compressed measurements and high-fidelity video recovery.

\begin{figure}[t]
\centering
\includegraphics[width=\columnwidth]{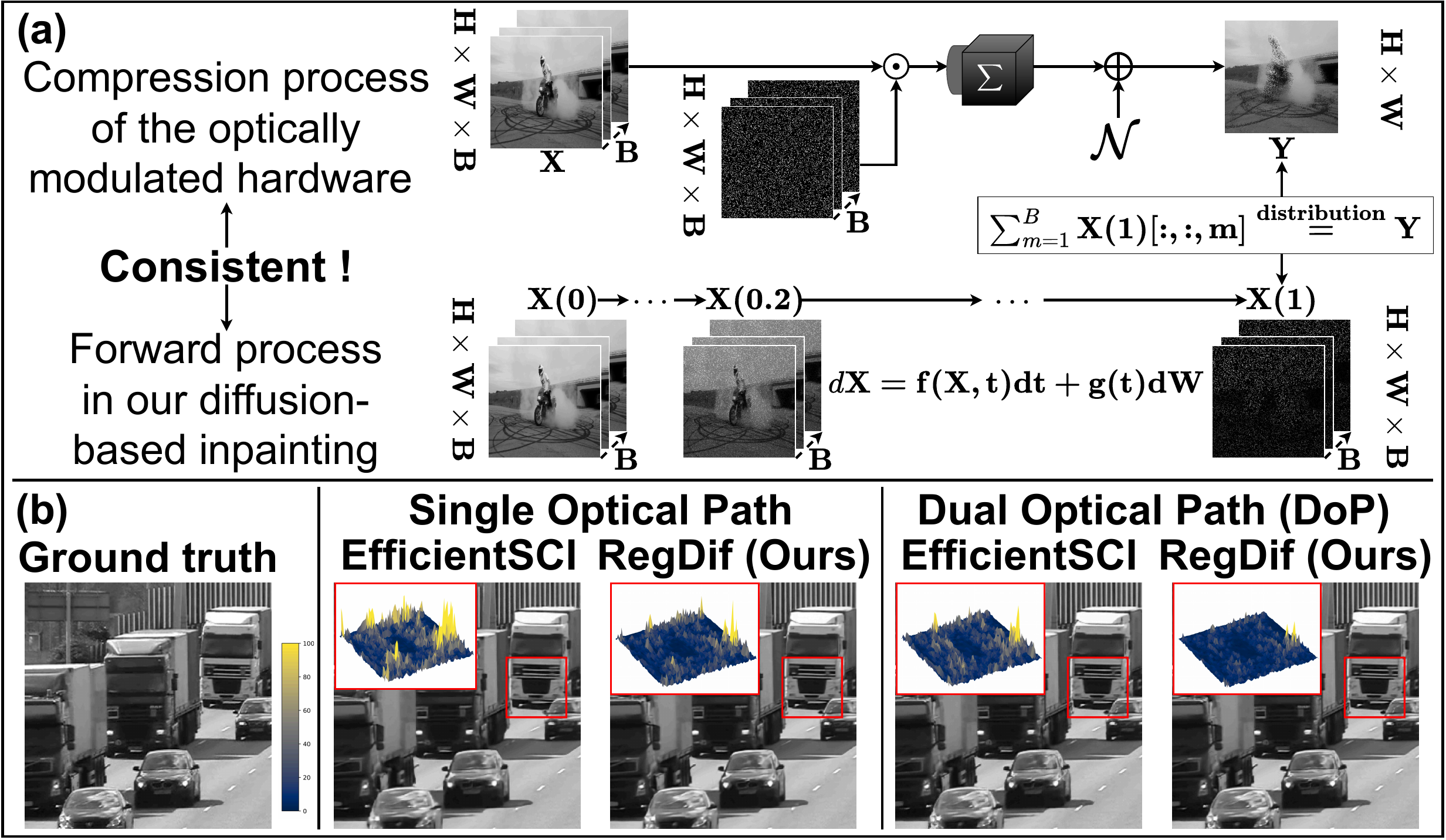} % Reduce the figure size so that it is slightly narrower than the column. Don't use precise values for figure width.This setup will avoid overfull boxes.
\caption{(a) The forward process in our diffusion-base inpainting aligns with the hardware compression process in video SCI. (b) Our method shows superiority in both single and dual-path settings. The 3D heatmap represents the absolute error between the red block and the ground truth.}
\label{fig:teasing}
\end{figure}

% The performance of video SCI primarily depends on the encoding mask and corresponding reconstruction algorithm. Mainstream SCI systems \cite{qiao2020snapshot,qiao2020deep} typically utilize random binary masks, which are physically implemented via digital micromirror devices (DMDs) for hardware encoding. Despite their success, two key limitations persist. \emph{Inefficient Light Utilization and Spatial Resolution Loss}: During the encoding process, the DMD reflects signals corresponding to mask values of 1 toward the detector, while signals associated with mask values of 0 are discarded, resulting in photon wastage and degradation of spatial resolution. \emph{Temporal Signals Superimposing and Dynamic Range Constraints}: Multiplexing of the encoding process introduces temporal information superimposing in the compressed measurements, complicating the reconstruction problem. Furthermore, due to the limited bit depth of the imaging sensor, the multiplexed signals must be regulated to avoid sensor saturation, thereby restricting the effective dynamic range of the measurement. As illustrated in Figure 2(a), the histogram of the measurements modulated by the random binary mask indicates the overlap of information from different frames, which also leads to a limited dynamic range. To address these issues, we propose a novel video SCI system differing in both hardware and software. 

\begin{figure}[t]
\centering
\includegraphics[width=0.96\columnwidth]{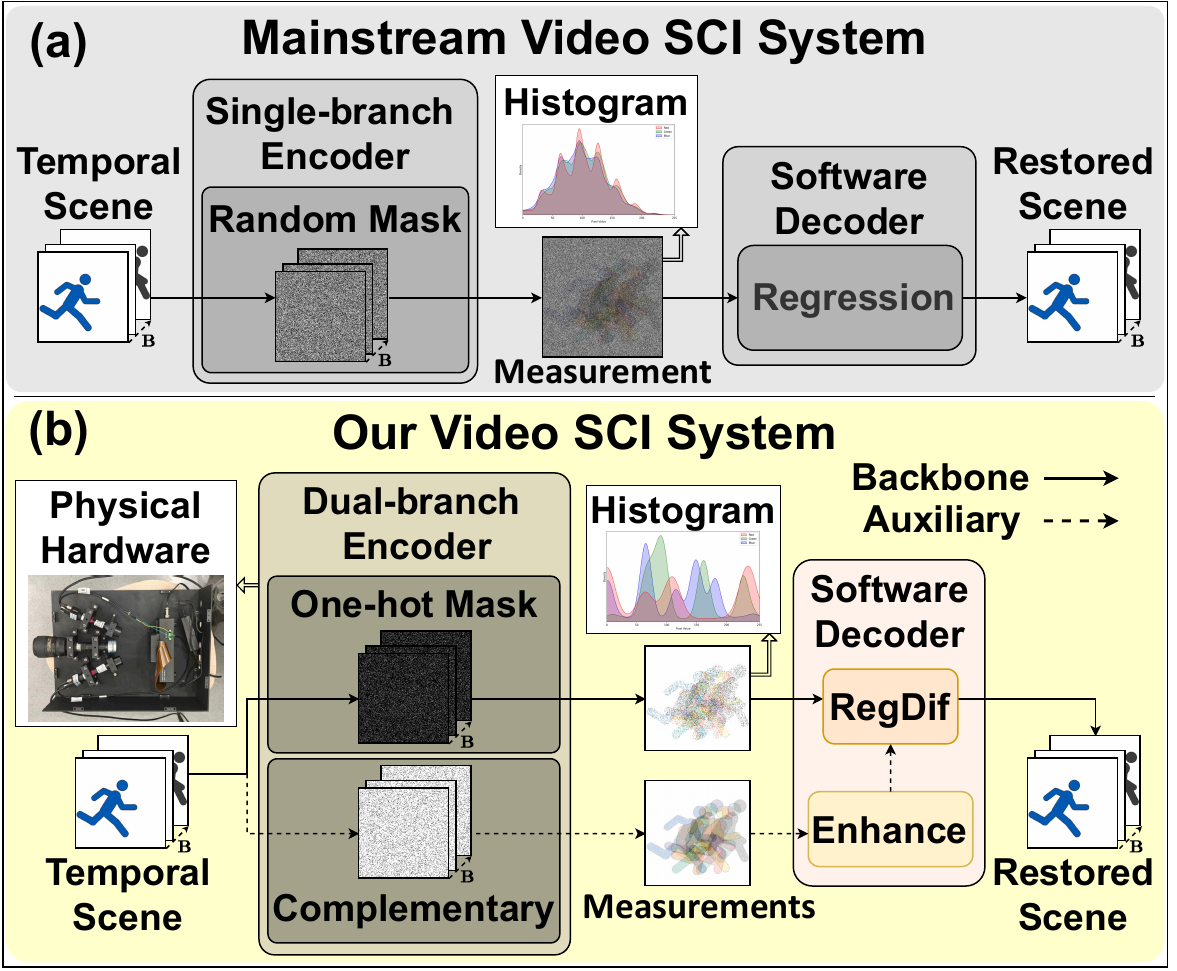} % Reduce the figure size so that it is slightly narrower than the column. Don't use precise values for figure width.This setup will avoid overfull boxes.
\caption{(a) Overall framework of a single-branch system using a random binary mask and its measurement pixel histogram. (b) Overall framework of a dual-branch system using a one-hot mask and its measurement pixel histogram.}
\label{fig:overall}
\end{figure}

The performance of video SCI primarily depends on the encoding mask and the corresponding reconstruction algorithm. Mainstream SCI systems \cite{qiao2020snapshot,qiao2020deep} typically employ random binary masks, which are physically implemented via digital micromirror devices (DMDs) for hardware encoding. Despite their success,  this approach inevitably results in
\emph{temporal aliasing during compression}. Specifically, multiplexing of the temporal channel in the encoding process leads to the superposition of information from different frames in the compressed measurement. As illustrated in Fig.~\ref{fig:overall}(a) histogram, different frames in the original scene exhibit distinct colors; however, after modulation by the random binary mask, all colors are blended in the measurement. Under the one-hot mask configuration, the DMD reflects only one sub-frame's signal per spatial location to the sensor during compression. This ensures perfectly decoupled temporal information, where each measurement pixel corresponds exclusively to a single original frame, thus eliminating temporal superposition. As shown in Fig.~\ref{fig:overall}(b) histogram, the measurement retains distinct inter-frame color separation. Furthermore, one-hot mask enables full-dynamic-range measurement capture, preventing bit-depth clipping and proving particularly advantageous in low-light and long-exposure scenarios, as well as in deploying a complete video SCI system on a chip.

Currently, no algorithms are specifically designed for one-hot masks to fully leverage their ability to decouple temporal information \cite{qiao2022coded}. To fill this gap, we reformulate reconstruction as a generative inpainting task and propose a video SCI-adapted stochastic differential equation (SDE) of the forward process that mirrors the hardware optically modulated encoding, as shown in Fig.~\ref{fig:teasing}(a). For our video SCI inpainting input, masked regions dominate each frame's spatial domain. To overcome degradation caused by large masked regions and avoid slow iterative sampling, we develop a hybrid framework dubbed RegDif, where a single regression block provides the coarse estimate, followed by the one-step diffusion for refinement. To our best knowledge, this represents the first application of a diffusion model to video SCI. Additionally, to further mitigate spatial degradation resulting from the one-hot mask, we introduce a dual optical path: the original path captures modulated signals whose corresponding one-hot mask values are 1, while a compensatory path with another 2D sensor records discarded signals with mask values of 0. A fusion module then enhances the inpainted video using the compensatory-path measurement. As shown in Fig.~\ref{fig:teasing}(b), selected frames of reconstructed videos show the superiority of our method.

In summary, we propose a novel video SCI system as shown in Fig.~\ref{fig:overall}(b), and the contributions of this work are:
\begin{itemize}
    \item Leveraging the temporal decoupling property of the one-hot mask, we transform the regression reconstruction problem into a diffusion-based inpainting task and propose a customized SDE of the forward process that aligns with the hardware compression process in video SCI.
    \item We observe that the pure diffusion method yields unsatisfactory reconstruction performance and speed in video SCI tasks characterized by severe degradation. To address these, we propose a hybrid framework employing one-step regression to obtain a coarse estimate, followed by one-step diffusion refinement. Additionally, we introduce a dual optical path to enhance the inpainted video.
    \item We conduct comprehensive experiments to demonstrate the exceptional performance of the proposed reconstruction framework tailored for the one-hot mask under both single and dual-optical-path configurations.
\end{itemize}

\section{Related Work}
\subsection{Video Snapshot Compressive Imaging}
In the hardware component of video SCI, a random binary mask is commonly implemented by either a digital micro-mirror device (DMD) \cite{qiao2020deep, qiao2020snapshot} or liquid crystal on silicon (LCOS) \cite{hitomi2011video, liu2013efficient}. To overcome issues caused by temporal aliasing, learned binary masks via programmable pixel sensors \cite{martel2020neural} and structural masks with reduced refresh rates \cite{wang2023deep} have been proposed. Additionally, optical designs that capture two objects in a single measurement \cite{qiao2020snapshot} and dual paths for complementary measurements of one object \cite{wang2025dual} have been explored. In this study, we are the first to integrate the one-hot mask with the dual optical path to decouple temporal information and improve light efficiency to mitigate the degradation issue.

In the reconstruction process, operating under default random binary mask settings, traditional model-based methods employ various regularizations, including Total Variation \cite{yuan2016generalized}, Gaussian Mixture Model \cite{yang2014compressive}, and Low Rank \cite{liu2018rank}. Plug-and-play (PnP) frameworks \cite{yuan2020plug, yuan2021plug} integrate pretrained deep image predictors into iterative optimizations for flexibility. For rapid reconstruction with high quality, deep learning-based methods \cite{cheng2020birnat, wang2023efficientsci, cheng2021memory, ma2019deep, wang2021metasci, wang2022spatial} have been introduced. BIRNAT \cite{cheng2020birnat} utilizes a bidirectional recurrent neural network for reconstruction, while RevSCI \cite{cheng2021memory} employs a 3D CNN-based memory-efficient framework. Additionally, STFormer \cite{wang2022spatial} and EfficientSCI \cite{wang2023efficientsci} leverage spatial-temporal transformers and spatial CNNs with temporal transformers, respectively. All previously mentioned deep learning-based methods depend on regression reconstruction. In contrast, we propose the first diffusion-based method tailored for the one-hot mask.

\subsection{Video Inpainting}
Video inpainting seeks to restore gaps or missing regions with visually consistent content while maintaining spatial and temporal coherence. Critical techniques include flow-based propagation \cite{gao2020flow,li2022towards,zhang2022flow,zhou2023propainter} and video transformers \cite{liu2021fuseformer,li2022towards,zhang2022flow}, with Propainter \cite{zhou2023propainter} integrating both in the architecture. With advancements in diffusion models, recent studies have explored the application of image diffusion models for video editing \cite{ceylan2023pix2video,geyer2023tokenflow,qi2023fatezero,wu2023tune}, such as AVID \cite{zhang2024avid} using motion modules and structure guidance. These methods emphasize the temporal continuity of masked objects. In contrast, our video SCI inpainting employs temporally incoherent one-hot masks. Inspired by IR-SDE \cite{luo2023image}, which is used for image restoration, we propose an SDE of the forward process that aligns with the video SCI encoding process and a tailored reconstruction method for severely degraded videos.

\section{Methodology}
% In this section, we first present the mathematical model of video SCI and extend it to our dual-path video SCI system with one-hot modulation. Subsequently, we introduce the forward and corresponding reverse SDEs for video SCI. We further analyze and elucidate several issues associated with pure diffusion methods in the video SCI and propose a hybrid reconstruction method that integrates regression and diffusion. Finally, we present a fusion method to integrate information from the compensatory path with the measurements from the active path. For clarity, all derivations assume single-channel grayscale input unless stated otherwise; color data processing is detailed in Appendix B.

% We first present the mathematical model video SCI and extend it to our dual-path system with one-hot mask. Then we transform the regression problem into a diffusion-based generation using the one-hot mask property and propose an SDE whose forward process matches the hardware compression. Furthermore, we introduce a hybrid method integrating regression and diffusion, addressing limitations of pure diffusion methods in video SCI. Finally, we show how to enhance the inpainted video using compensatory-path measurements. For clarity, all derivations assume grayscale input; color processing is shown in Appendix B.

\subsection{Mathematical Model of Video SCI}
All subsequent discussions will focus on grayscale videos, while color processing is illustrated in Appendix D.

As shown in Fig. 2(a), in the single-path hardware encoder with random binary mask, the original $B$ frame input video $\Xmat \in \mathbb{R}^{H \times W \times B}$ is modulated by the mask $\tilde{\Mmat} \in \mathbb{R}^{H \times W \times B} \sim \text{Bernoulli}(0.5)^{H \times W \times B}$. Subsequently, by compressing data across time, the 2D camera detector captures a compressed measurement $\tilde{\Ymat} \in \mathbb{R}^{H \times W}$. The entire encoding process can be represented as follows:
\begin{equation*}
    \tilde{\Ymat} = \textstyle \sum_{m=1}^B \Xmat_m \odot \tilde{\Mmat}_m + \tilde{\Zmat},
    \label{eq:encoding}
\end{equation*}
where $\odot$ denotes the Hadamard (element-wise) multiplication. The notation $\Xmat_m \mathop{:=} \Xmat[:, :, m]$ represents the $m$-th frame of the input video, while $\tilde{\Mmat}_m \mathop{:=} \tilde{\Mmat}[:, :, m]$ indicates the corresponding modulation for the $m$-th frame of the video. Additionally, $\tilde{\Zmat} \in \mathbb{R}^{H \times W}$ denotes the measurement noise. Subsequently, a regression-based reconstruction algorithm, denoted as $\mathcal{D}^*_{\text{reg}} \colon \mathbb{R}^{H \times W} \times 
\mathbb{R}^{H \times W \times B} \to \mathbb{R}^{H \times W \times B}$, is employed to recover the original video, determined by:
\begin{equation*}
    \mathcal{D}_{\text{reg}}^* = \textstyle \argmin_{\mathcal{D}_{\text{reg}}} || \Xmat - \mathcal{D}_{\text{reg}}(\tilde{\Ymat}, \tilde{\Mmat})||_F.
\end{equation*}

In our encoding, the one-hot mask is generated through:
\begin{equation}
    \Mmat[h,w,m] = 
\begin{cases}
    1, & \text{if } \left\lfloor \mathcal{I}[h,w] \right\rfloor = m, \\
    0, & \text{otherwise},
\end{cases}
\label{eq:one_hot_mask}
\end{equation}
where $\lfloor \cdot \rfloor$ denotes the floor function, which rounds down to the nearest integer, and $\mathcal{I} \sim \text{Uniform}(1,B+1)^{H \times W}$. According to the mask defined in Eq.~\eqref{eq:one_hot_mask}, at any spatial position $(h, w)$, only one temporal channel has a value of 1 while all others are 0, which is why we refer to it as a one-hot mask.

In addition to the one-hot mask $\Mmat$, our encoder incorporates a compensatory path alongside the original path to capture signals associated with mask values of 0. Our optical design enables the measurement $\Ymat^C$ from the compensatory path to capture signals that are not detected by the measurement $\Ymat$ from the original path, thereby positioning $\Ymat^C$ as a complement to $\Ymat$. Consequently, the encoding process in our dual optical path system is represented as follows:
\begin{eqnarray}
     \Ymat &=& \textstyle \sum_{m=1}^B \Xmat_m \odot \Mmat_m + \Zmat,
    \label{eq:y_active}\\
    \Ymat^C &=&  \textstyle \sum_{m=1}^{B} \Xmat_m \odot (\mathbf{1}-\Mmat_m) + \Zmat^C,
    \label{eq:y_compensatory}
\end{eqnarray}
where $\Zmat$/$\Zmat^C$ is the corresponding measurement noise in the original/compensatory path, and $\mathbf{1}\in \mathbb{R}^{H \times W}$ is an all-one matrix. More hardware encoder details are in Appendix A.

We observe that using one-hot mask $\Mmat$ can separate the temporal information of different frames within the measurement $\Ymat$ from Eq.~\eqref{eq:y_active}. For instance, the non-zero components of $\Ymat \odot \Mmat_m$ represent the pixel values at the corresponding positions in the $m$-th frame of the input video with the measurement noise (commonly modeled as a Gaussian distribution). Consequently, it is apt to employ diffusion-based inpainting for reconstruction. Intuitively, we consider the following SDE of the forward process in diffusion:
\begin{equation}
    d\Xmat=\underbrace{\mu(t)(\Xmat_{\text{dst}}-\Xmat(t))dt}_{\text{drift}}+\underbrace{\sigma(t)d\Wmat}_{\text{dispersion}},
    \label{eq:forward_SDE}
\end{equation}
with the boundary condition constraint $\Xmat_{\text{src}} \mathop{:=}\Xmat(0)=\Xmat$ and $t\in[0,1]$, where $\mu(t), \sigma(t)$ are time-dependent hyperparameters that characterize speed and stochastic volatility, and $\Xmat_{\text{dst}}=\Xmat\odot\Mmat$, where $\Wmat$ represents the Wiener process. It can be observed that the drift term is analogous to the interpolation method that gradually transforms $\Xmat_{\text{src}}$ into $\Xmat_{\text{dst}}$. Additionally, the role of the dispersion term is to introduce Gaussian noise during this process, serving a similar function as the measurement noise $\Zmat$ in Eq.~\eqref{eq:y_active}. Fortunately, our investigation reveals that similar SDE methods have been thoroughly examined in prior research \cite{luo2023image}, which focused on image restoration. Consistent with prior research, we discover that if $\mu(t)$ and $\sigma(t)$ are constrained by $\sigma^2(t)/\mu(t)=2\lambda^2$, where $\lambda^2$ is the stationary variance, Eq.~\eqref{eq:forward_SDE} yields a closed-form solution:
\begin{equation}
    \Xmat(t) = {\left[(1-\bar{\mu}(t))\Xmat_{\text{src}}+\bar{\mu}(t)\Xmat_{\text{dst}}\right]} + {\bar{\sigma}(t)\epsilon},
    \label{eq:forward}
\end{equation}
where $\bar{\mu}(t) = 1 - e^{-\theta(t)}$ is the interpolation factor, $\bar{\sigma}(t)=\lambda\sqrt{1-e^{-2\theta(t)}}$, $\theta(t)=\int_0^t \mu(s)ds$ , and $\epsilon \sim \mathcal{N}(0,1)^{H \times W \times B}$. By appropriately configuring $\lambda$ and $\mu(t)$ (where $\sigma(t)$ can be determined by $\lambda\sqrt{2\mu(t)}$), we can achieve the following approximations: as $t$ progresses from $0$ to $1$, $\bar{\mu}(t)$ transitions from $0$ to $1$, and $\bar{\sigma}(t)$ transitions from $0$ to $\lambda$. This configuration ensures that the diffusion forward process aligns with the video SCI hardware encoding process in Eq.~\eqref{eq:y_active}. Specifically, it is easy to notice that $\Xmat(0)$ is consistent with the input video $\Xmat$. Assuming the measurement noise $\Zmat \sim \mathcal{N}(0, \tilde{\lambda}^2)$ (for simplicity, we consider the case where the mean of the noise is zero; in scenarios where a non-zero mean is considered, a modification to $\Xmat_{\text{dst}}$ is required), when we set $\lambda = \tilde{\lambda}/\sqrt{B}$, compressed result from $\Xmat(1)$ alongside the temporal channel is consistent with $\Ymat$, such that $\sum_{m=1}^B\Xmat(1)[:,:,m]\overset{\text{distribution}}{=}\Ymat$. Thus, we can characterize the process of compressing the original video $\Xmat$ into $\Ymat$ using the video SCI hardware encoder via the diffusion forward process from $\Xmat(0)$ to $\Xmat(1)$. A more formal statement and the detailed proof are shown in Appendix B.

\begin{figure}[t]
\centering
\includegraphics[width=0.95\columnwidth]{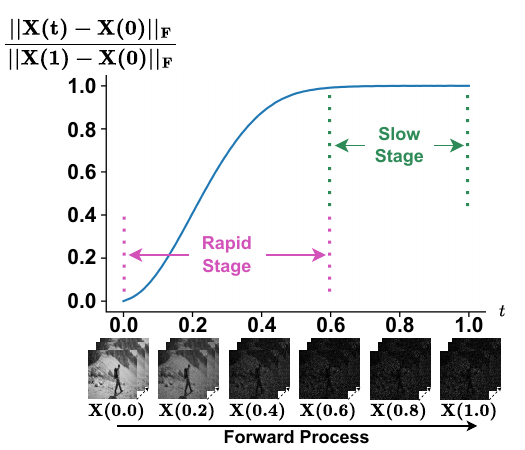} % Reduce the figure size so that it is slightly narrower than the column. Don't use precise values for figure width.This setup will avoid overfull boxes.
\caption{The normalized $\ell_2$-norm distance between $\Xmat(t)$ and $\Xmat(0)$ as t varies from 0 to 1 in our forward process.}
\label{fig:nonlinear}
\end{figure}

\begin{algorithm}[t]
\caption{Pure Diffusion-based Inpainting $\mathcal{D}_{\text{dif}}^*$}
\label{alg:dif}
\textbf{Input}: Measurement $\Ymat\in \mathbb{R}^{{H\times W}}$, Mask $\Mmat\in\mathbb{R}^{{H\times W \times B}}$\\
\textbf{Parameter}: Noise Predictor $\epsilon_{\gamma^*}(\cdot,\cdot)$, Iteration Count $N$, Station Variance $\lambda^2$, Discrete Scheduler $\mu=[\mu_1, \dots, \mu_N]$\\
\textbf{Output}: Reconstructed Video $\Xmat$
\begin{algorithmic}[1] %[1] enables line numbers
% \FOR{$t=T$ \DOWNTO $1$}
\STATE $\Delta t \leftarrow 1/N$
\STATE $H,W,B \leftarrow \text{shape}(\Mmat)$
\STATE $\sigma \leftarrow \lambda\sqrt{2\mu}$
\STATE $\theta \leftarrow \text{cumsum}(\mu\cdot \Delta t)$
\STATE $\bar{\sigma} \leftarrow \lambda\sqrt{1-e^{-2\theta}}$
\FOR{$m=1,\dots,B$}
\STATE $\tilde{\Xmat}[:,:,m] \leftarrow Y \odot M[:,:,m]$ 
\ENDFOR
\STATE $\Xmat \leftarrow \tilde{\Xmat}$
\FOR{$i=N,\dots,1$}
\STATE $\hat{\epsilon} \leftarrow \mathcal{N}(0,\Delta t)^{H\times W \times B}$
\STATE $\epsilon \leftarrow \epsilon_{\gamma^*}(\Xmat, i\cdot\Delta t)$
\STATE $(\Delta \Xmat)_{\epsilon} \leftarrow \left(\mu[i](\tilde{\Xmat}-\Xmat)+\frac{\sigma[i]^2}{\bar{\sigma}[i]} \epsilon\right)\Delta t+\sigma[i]\hat{\epsilon}$
\STATE $\Xmat \leftarrow \Xmat - (\Delta\Xmat)_{\epsilon}$
\ENDFOR
\STATE \textbf{return} $\Xmat$
\end{algorithmic}
\end{algorithm}

We can then reverse the process to restore $X(0)$ from $X(1)$ by backing the SDE in time \cite{song2020score}. The corresponding SDE of the reverse process is given by:
\begin{equation}
    % \begin{aligned}
        (d\Xmat)_{\epsilon_t} = [\mu(t)(\Xmat_{\text{dst}}-\Xmat(t))+\tilde{\sigma}(t)\epsilon_t]dt+\sigma(t)d\hat{\Wmat},
        % &\approx [\mu(t)(\Xmat(T)-\Xmat(t))+\tilde{\sigma}(t)\epsilon_t]dt+\sigma(t)d\hat{\Wmat}
        \label{eq:backward_SDE}
    % \end{aligned}
\end{equation}
where $\tilde{\sigma}(t) = \sigma(t)^2/\bar{\sigma}(t)$, $\hat{\Wmat}$ is a reverse-time Wiener process, and $\epsilon_t$ is the noise we need to approximate by a neural network $\epsilon_{\gamma^*}(\Xmat(t), t)$, determined by the trajectory loss:
\begin{equation}
    \scalebox{0.95}{$\gamma^* = \textstyle \argmin_{\gamma} \mathbb{E}_{\scriptscriptstyle{t\sim U(0,1)}} [ ||(d\Xmat)_t^*-(d\Xmat)_{\epsilon_{\gamma}(\Xmat(t), t)}||_F ]$},
    \label{eq:trajectory_loss}
\end{equation}
% where $(d\Xmat)^*_t$ denotes the ideal trajectory between $t-dt$ and $t$. In contrast to the commonly used objective function in DDPM \cite{ho2020denoising}, trajectory loss is employed to stabilize training when restoring images affected by complex degradations \cite{luo2023image}. In this study, we retain this trajectory loss because the degradation issues in our video inpainting task are more pronounced than those encountered in typical image restoration tasks. After deriving the trained noise predictor $ \epsilon_{\gamma^*}(\cdot, \cdot) $, one can retrieve the original video through iteratively denoising from $\Xmat_{\text{dst}}$ by applying Eq.~\eqref{eq:backward_SDE}. It is noteworthy that in the practical reverse process, we substitute $\Xmat_{\text{dst}}[:,:,m]$, which is unknown, with $\Ymat \odot M_m$ for $m=1,\dots,B$, where the error is a higher-order infinitesimal in comparison to $d\hat{\Wmat}$. Further elaborations about our inpainting, such as how to derive Eq.~\eqref{eq:forward} from Eq.~\eqref{eq:forward_SDE}, the acquisition of Eq.~\eqref{eq:backward_SDE}, the details of Eq.~\eqref{eq:trajectory_loss}, and a detailed algorithm are provided in Appendix B.
where $(d\Xmat)^*_t$ denotes the ideal trajectory between $t-dt$ and $t$. In contrast to the commonly used objective function in DDPM \cite{ho2020denoising}, trajectory loss is employed to stabilize training when restoring images affected by complex degradations \cite{luo2023image}. In this study, we retain trajectory loss as degradation in our video inpainting task is more severe than in typical image restoration tasks. After deriving the trained noise predictor $ \epsilon_{\gamma^*}(\cdot, \cdot) $, one can use $\mathcal{D}^*_{\text{reg}}$ to retrieve the original video, as shown in Algorithm \ref{alg:dif}. It is noteworthy that in the practical reverse process, we substitute $\Xmat_{\text{dst}}[:,:,m]$, which is unknown, with $\Ymat \odot \Mmat_m$ for $m=1,\dots,B$, where the error is a higher-order infinitesimal compared with $d\hat{\Wmat}$. Further details regarding our diffusion-based inpainting, including the derivation of Eq.~\eqref{eq:forward} from Eq.~\eqref{eq:forward_SDE}, the acquisition of Eq.~\eqref{eq:backward_SDE}, and the specifics of Eq.~\eqref{eq:trajectory_loss}, can be found in Appendix B.

\subsection{RegDif: One-step Regression + One-step Diffusion}
\begin{algorithm}[t]
\caption{Regress Init + Diffusion Refine $\mathcal{D}_{\text{regdif}}^*$}
\label{alg:regdif}
\textbf{Input}: Measurement $\Ymat\in \mathbb{R}^{H\times W}$, Mask $\Mmat\in \mathbb{R}^{H\times W\times B}$\\
\textbf{Parameter}: Regressor Initializer $G_{\alpha^*}(\cdot)$, Timestep Predictor $H_{\beta^*}(\cdot)$, Noise Predictor $\epsilon_{\gamma^*}(\cdot,\cdot)$, Station Variance $\lambda^2$, Continuous Scheduler $\mu(\cdot)$\\
\textbf{Output}: Reconstructed Video $\Xmat$
\begin{algorithmic}[1] %[1] enables line numbers
% \FOR{$t=T$ \DOWNTO $1$}
\STATE $H,W,B \leftarrow \text{shape}(\Mmat)$
\STATE $\sigma(\cdot) \leftarrow \lambda\sqrt{2\mu(\cdot)}$
\STATE $\theta(\cdot) \leftarrow \int_0^{\cdot}(\mu(s)ds$
\STATE $\bar{\sigma}(\cdot) \leftarrow \lambda\sqrt{1-e^{-2\theta(\cdot)}}$
\FOR{$m=1,\dots,B$}
\STATE ${\Xmat}(1)[:,:,m] \leftarrow Y \odot M[:,:,m]$
\ENDFOR
\STATE $\hat{\Xmat}(\hat{t}) \leftarrow G_{\alpha^*}(\Xmat(1))$
\STATE $\bar{\Xmat}(\hat{t}) \leftarrow \Mmat\odot \Xmat(1)+(\mathbf{1}-\Mmat)\odot \hat{\Xmat}(\hat{t})$
\STATE $\hat{t} \leftarrow H_{\beta^*}(\bar{\Xmat}(\hat{t}))$
\STATE $\hat{\epsilon} \leftarrow \mathcal{N}(0, \hat{t})^{H \times W \times B}$
\STATE $\epsilon \leftarrow \epsilon_{\gamma^*}(\bar{\Xmat}(\hat{t}), \hat{t})$
\STATE $(\Delta \Xmat)_{\epsilon} \leftarrow \left(\mu(\hat{t})(\Xmat(1)-\bar{\Xmat}(\hat{t}))+\frac{\sigma(\hat{t})^2}{\bar{\sigma}(\hat{t})} \epsilon\right)\hat{t}+\sigma(\hat{t})\hat{\epsilon}$
\STATE $\Xmat \leftarrow \bar{\Xmat}(\hat{t}) - (\Delta\Xmat)_{\epsilon}$
\STATE \textbf{return} $\Xmat$
\end{algorithmic}
\end{algorithm}

\begin{figure*}[t]
\centering
\includegraphics[width = 0.96\textwidth]{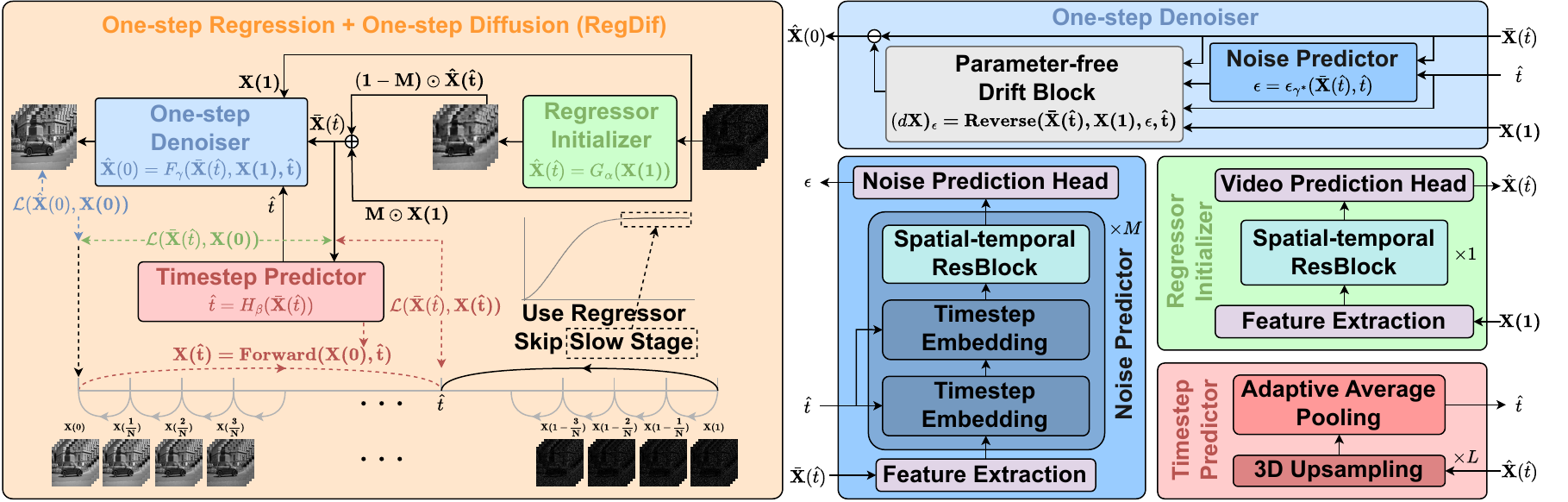} % Reduce the figure size so that it is slightly narrower than the column. Don't use precise values for figure width.This setup will avoid overfull boxes.
\captionsetup{width=\linewidth}
\caption{Illustration of the overall RegDif framework (LHS) and key components in RegDif (RHS). In the LHS, black solid line indicates the pipeline of RegDif, gray solid line indicates the reverse process of pure diffusion inpainting. Three colored dash lines indicate pipelines for different loss derivations. The RHS shows architectures of key components mentioned in RegDif.}
\label{fig:software_framework}
\end{figure*}

% Illustration of overall RegDif framework (LHS) and key components in RegDif (RHS). In LHS RegDif framework, black solid line indicates the pipeline of RegDif, gray solid line indicates reverse process of pure diffusion. Three different colored lines indicate the pipeline of different loss derivations. RHS shows the architectures of key components mentioned in RegDif framework.

In executing the pure diffusion-based inpainting task described above, we find that the iterative process limits rapid reconstruction and yields suboptimal performance. Our analysis attributes these unsatisfactory reconstruction results to the nonlinear characteristics of the interpolation factor $\bar\mu(t)$ in Eq.~\eqref{eq:forward}. For instance, in the 100-step diffusion with a simple linear schedule, as illustrated in Fig.~\ref{fig:nonlinear}, we note that after approximately 60 iterations in the forward process, the difference between the current state $\Xmat(0.6)$ and the terminal state $\Xmat(1)$ becomes negligible. Consequently, during the reverse process, the model learns inconsistent denoising strategies: smooth in the slow stage and fast in the rapid stage. Additionally, the degradation of $\Xmat(1)$ caused by the one-hot mask further constrains denoising.

To address this, one solution is to design an intricate scheduler $\mu(t)$ with parameter tuning. Instead, we use a simple regressor with one block to replace the slow reverse stage, followed by one-step diffusion denoising. Based on our design, this approach ($i$) significantly enhances reconstruction speed, ($ii$) eliminates the need for complex parameter tuning, and ($iii$) partially mitigates degradation of the masked video via the regressor's coarse prediction.

% As illustrated in Fig.~\ref{fig:software_framework}, our framework comprises three key components: the Regressor Initializer, the Timestep Predictor, and the One-step Denoiser. Initially, we employ the Regressor Initializer to smooth the degraded masked video, which has been derived by decoupling temporal information from measurement $\Ymat$. Subsequently, we use the coarse prediction result to update the masked regions in the degraded video. Next, we utilize the Timestep Predictor to forecast the timestep of the updated video, after which we apply the One-step Denoiser to conduct denoising as described in Eq.~\eqref{eq:backward_SDE}.

% As shown in Fig.~\ref{fig:software_framework}, our framework has three key components: Regressor Initializer, Timestep Predictor, and One-step Denoiser. First, we use the Regressor Initializer to smooth the degraded masked video obtained by decoupling temporal information from measurement $\Ymat$. Then we update the masked regions using this coarse prediction. Next, the timestep of the updated video is derived from Timestep Predictor, after which the One-step Denoiser performs denoising per Eq.~\eqref{eq:backward_SDE}.

As shown in Fig.~\ref{fig:software_framework}, our framework consists of three components: Regressor Initializer, Timestep Predictor, and One-step Denoiser. The Regressor Initializer first smooths the degraded masked video obtained by decoupling temporal information from measurement $\Ymat$ through a {\em one-hot mask} and updates masked regions using this coarse prediction. Then, the Timestep Predictor derives the timestep of the updated video. After that, the {\em one-step denoiser} directly denoises the updated video into the final result via Eq.~\ref{eq:backward_SDE} without iteration. 

% Then, the Timestep Predictor derives the timestep of the updated video, followed by denoising from the {\em one-step denoiser}.

\begin{figure}[t]
\centering
\includegraphics[width=0.96\columnwidth]{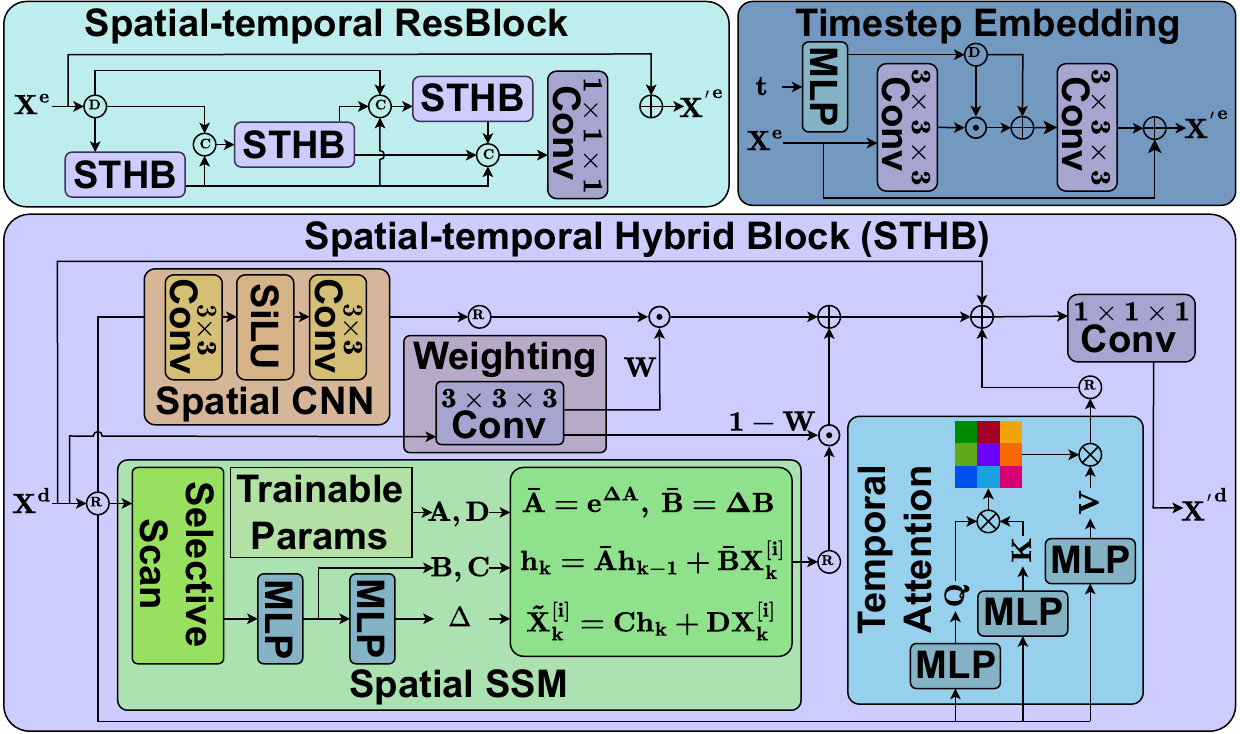} % Reduce the figure size so that it is slightly narrower than the column. Don't use precise values for figure width.This setup will avoid overfull boxes.
\caption{Detailed architectures of the Spatial-temporal ResBlock, the STHB block, and the Timestep Embedding block.}
\label{fig:software_detail}
\end{figure}

% In the aforementioned framework, we employ 3D Convolution with Average Pooling in the Timestep Predictor to derive the corresponding timestep for the updated video. For Feature Extraction and Prediction Head blocks in the Noise Predictor and Regressor Initializer, we utilize identical 3D Convolution structures for feature embedding and final prediction. Moreover, the Noise Predictor and Regressor Initializer share the same Spatial-temporal ResBlock backbone; the key distinction is that the Noise Predictor incorporates Timestep Embedding blocks to enrich the input with its corresponding timestep. The Regressor Initializer employs only one Spatial-temporal ResBlock, thus referred to as one-step regression. In the Spatial-temporal ResBlock, we first partition the embedded inputs along the channel dimension then feed them into STHB blocks for processing, followed by the fusion of the processed results. Within each STHB, we leverage 2D Convolution and 2D State Space Model (SSM) to capture local and global information from each 2D frame while utilizing Attention to extract temporal information among different frames, as shown in Fig.~\ref{fig:software_detail}. Detailed explanations of the model structure can be found in Appendix C. 

In the Timestep Predictor, we use 3D Convolution with Average Pooling to obtain the current timestep. The Noise Predictor and Regressor Initializer share identical 3D Convolution structures for Feature Extraction and Prediction Head blocks, and both use the same Spatial-temporal ResBlock backbone, differing only in that the Noise Predictor adds Timestep Embedding blocks to predict time-dependent noise. The Regressor Initializer uses only one Spatial-temporal ResBlock, thus referred to as {\em one-step regression}. Within Spatial-temporal ResBlock, embedded inputs are partitioned along channels, processed through STHB blocks, and fused. Each STHB uses 2D Convolution and 2D State Space Model (SSM) for per-frame local/global information, and Attention for cross-frame temporal information, as shown in Fig.~\ref{fig:software_detail}. Model details are shown in Appendix C.

% Regarding the spatial-temporal ResBlock architecture, building on \cite{wang2023efficientsci}, which utilizes 2D Spatial Convolution and Temporal Attention, we introduce an additional 2D State Space Model (SSM) to capture long-range pixel relations and mitigate degradation issues. Detailed model structure can be found in Appendix C.

In Fig.~\ref{fig:software_framework}, we present three different loss terms that are designed for the joint training of $G_{\alpha}$, $H_{\beta}$, and $F_{\gamma}$:
\begin{equation*}
    \mathcal{L} = \underbrace{\mathcal{L}(\bar{\Xmat}(\hat{t}),\Xmat(0))}_{\text{Regression Loss}}+\underbrace{\mathcal{L}(\bar{\Xmat}(\hat{t}),\Xmat(\hat{t}))}_{\text{Alignment Loss}}+\underbrace{\mathcal{L}(\hat{\Xmat}(0),\Xmat(0))}_{\text{Diffusion Loss}},
\end{equation*}
where $\mathcal{L}(\cdot,\cdot)$ is the $\ell_2$-norm loss, $\bar{\Xmat}(\hat{t})$ is the video updated by the Regressor Initializer, $\Xmat(\hat{t})$ is the intermediate state by setting $t=\hat{t}$ in Eq.~\eqref{eq:forward}, $\hat{\Xmat}(0)$ is the result derived from the One-step Denoiser. Regression Loss ensures the coarse prediction result of the Regressor Initializer being closer to the ground truth, providing improved initialization input for the One-step Denoiser. Alignment Loss aligns the updated video within our reverse denoising process, allowing the application of Eq.~\eqref{eq:backward_SDE} for denoising. Finally, Diffusion Loss, a special case of trajectory loss in Eq.~\eqref{eq:trajectory_loss} by setting $(d\Xmat)^*_{\hat{t}}=\bar{\Xmat}(\hat{t})-\Xmat(0)$ and $\hat{\Xmat}(0)=\bar{\Xmat}(\hat{t})-(d\Xmat)_{\epsilon_{\hat{t}}}$, optimizes the parameters of the Noise Predictor in the One-step Denoiser, enabling one-step diffusion from $\bar{\Xmat}(\hat{t})$ to $\Xmat(0)$. With trained parameters $\alpha^*$, $\beta^*$, and $\gamma^*$, the algorithm $ \mathcal{D}_{\text{regdif}}^* $ retrieves the original video, as demonstrated in Algorithm \ref{alg:regdif}.

\begin{figure*}[t]
\centering
\includegraphics[width = 0.773\textwidth]{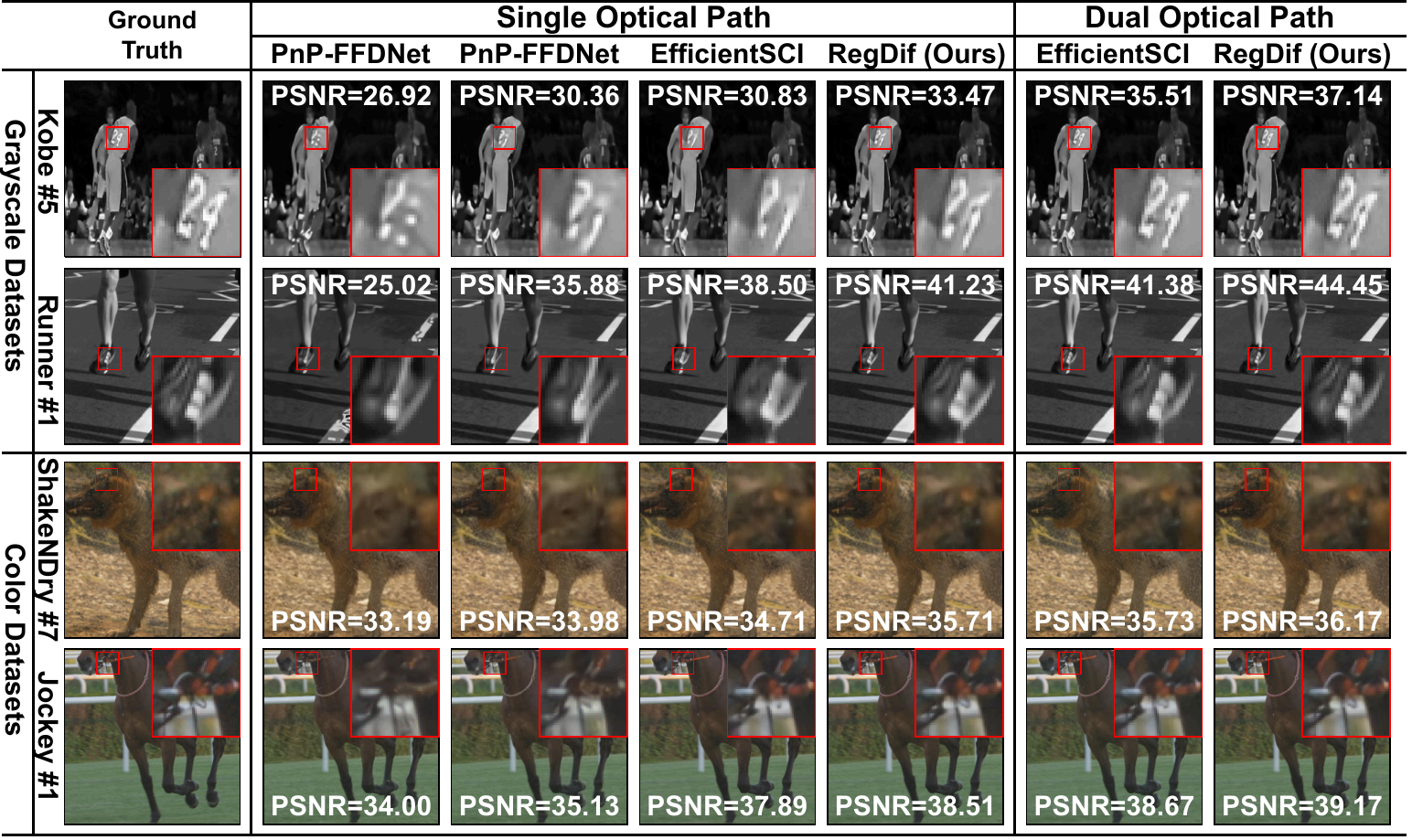} % Reduce the figure size so that it is slightly narrower than the column. Don't use precise values for figure width.This setup will avoid overfull boxes.
\caption{Selected reconstruction frames of simulated grayscale and color datasets. Zoom in for better view.}
\label{fig:sim_result}
\end{figure*}

\begin{figure}[t]
\centering
\includegraphics[width = 0.84\columnwidth]{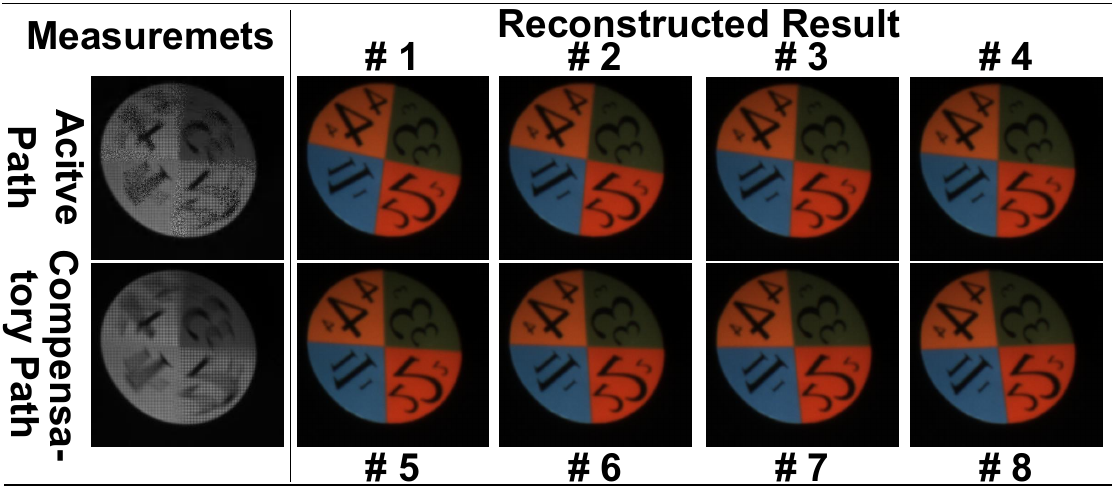} % Reduce the figure size so that it is slightly narrower than the column. Don't use precise values for figure width.This setup will avoid overfull boxes.
\caption{Measurements captured by our dual-path hardware with one-hot mask and reconstructed results of DoP-RegDif.}
\label{fig:real_result}
\end{figure}

Furthermore, drawing inspiration from the adapter, we integrate the compensatory measurement from Eq.~\eqref{eq:y_compensatory} into our reconstruction framework to enhance reconstruction quality, utilizing the architecture analogous to that of the adapter:
\begin{equation*}
    \begin{aligned}
        \hat{\mathbf{X}}^{\text{enhance}}(\hat{t}) &= \text{Fusion}_{x}\left(\hat{\mathbf{X}}(\hat{t}), \text{Embedding}_x(\mathbf{Y}^C)\right), \\
        \epsilon_{\hat{t}}^{\text{enhance}} &= \text{Fusion}_{\epsilon}\left(\epsilon_{\hat{t}}, \text{Embedding}_{\epsilon}(\mathbf{Y}^C)\right),
    \end{aligned}
\end{equation*}
where $\hat{\mathbf{X}}^{\text{enhance}}(\hat{t})$ and $\epsilon_{\hat{t}}^{\text{enhance}}$ denote the enhanced outputs of the Regressor Initializer and the Noise Predictor, respectively. The modules $\text{Embedding}_x$ and $\text{Embedding}_{\epsilon}$ serve to obtain the hidden representations of the measurement $\Ymat^C$, which are then fed into the video fusion module $\text{Fusion}_x$ and the noise fusion module $\text{Fusion}_{\epsilon}$, respectively. Details of the fusion architecture can be found in Appendix C.

% \xin{I am here!}

\section{Experiments}

\subsection{Datasets and Implementation Details}
Following the settings in previous works \cite{wang2023efficientsci, cheng2020birnat}, we use DAVIS2017 \cite{pont20172017} with resolution $480\times 894$ (480P) as the training dataset for the model. In the evaluation stage, we test the RegDif on six simulation grayscale/color datasets with a size of \{$256\times 256 \times 8$, $512 \times 512 \times 3 \times 8$\}. Subsequently, we conduct data acquisition and reconstruction in real-world scenarios using the dual optical path equipped with the one-hot mask.
For the reconstruction framework RegDif, as shown in Fig.~\ref{fig:software_framework}, we set 7 building blocks in the Noiser Predictor and 1 building block in the Regressor Initializer to match 8 building blocks commonly used in previous works \cite{wang2023efficientsci}. We use the PyTorch framework with 4 NVIDIA RTX 3090
GPUs for training with Adam optimizer ($\beta_1=0.9,\beta_2=0.99$) on a $1\times10^{-4}$ learning rate over 100 epochs with a decay ratio $0.5$ per 25 epochs and then finetune with a $1\times 10^{-5}$ learning ratio over 20 epochs. The Peak Signal to Noise Ratio (PSNR) and Structural SIMilarity (SSIM) \cite{wang2004image} are used as the performance indicators of the reconstruction quality. In all experiments, the best and second-best results of the evaluated methods are \textbf{highlighted} and \underline {underlined}.

\begin{figure}[t]
\centering
\includegraphics[width = 0.92\columnwidth]{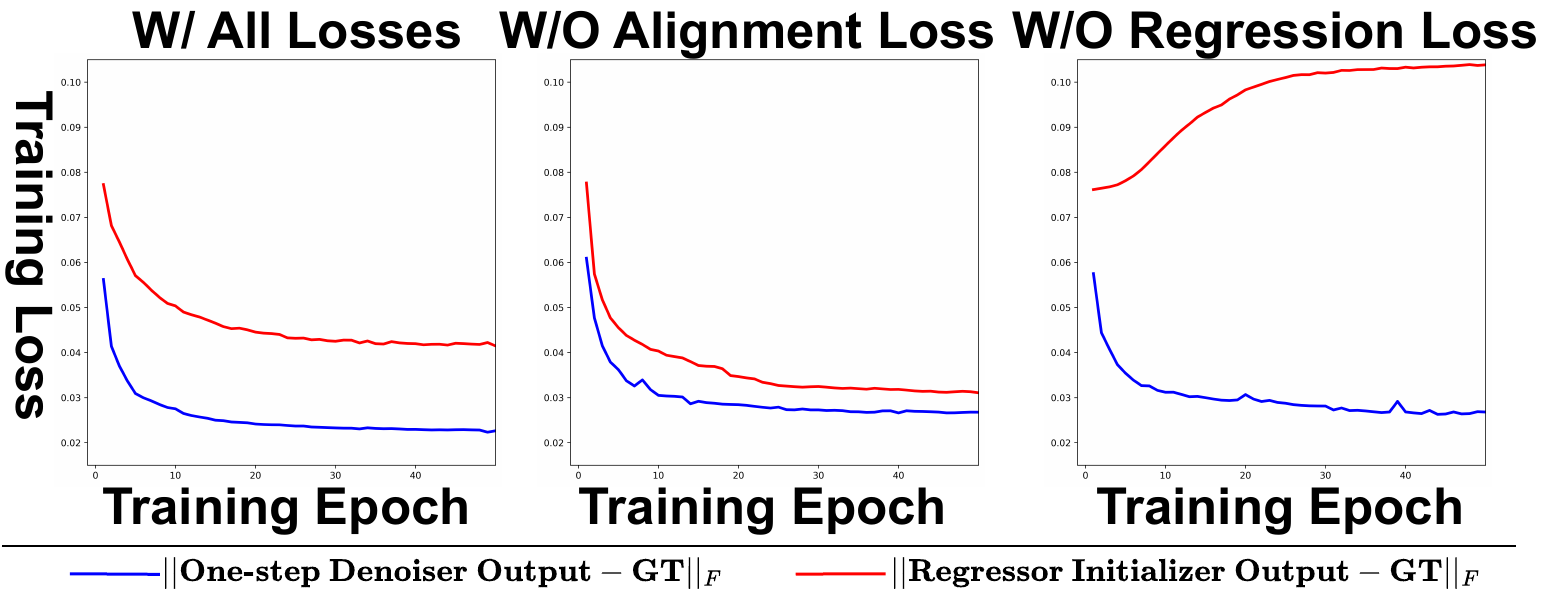} % Reduce the figure size so that it is slightly narrower than the column. Don't use precise values for figure width.This setup will avoid overfull boxes.
\caption{Ablation on the loss functions used in RegDif.}
\label{fig:ablation_loss}
\end{figure}

\subsection{Results on Simulation Benchmark Videos}

\begin{table*}
\centering
\begin{adjustbox}{width=0.955\textwidth}
\begin{tabular}{l|c|c|c|c|c|c|c} 
\hline
\multicolumn{1}{c|}{Grayscale Datasets} & Kobe                           & Traffic                        & Runner                         & Drop                           & Crash                          & Aerial                         & Average                         \\ 
\hline\hline
GAP-TV                            & 22.42, 0.690                   & 19.53, 0.627                   & 27.44, 0.861                   & 31.50, 0.932                   & 24.01, 0.836                   & 25.18, 0.828                   & 25.01, 0.796                    \\
PnP-FFDNet                        & 26.92, 0.912                   & 17.01, 0.694                   & 25.02, 0.902                   & 10.05, 0.341                   & 14.91, 0.539                   & 21.69, 0.798                   & 19.27, 0.698                    \\
PnP-FastDVDnet                    & 30.36, 0.937                   & 26.80, 0.920                   & 35.88, 0.957                   & 42.08, 0.989                   & 26.23, 0.906                   & 27.25, 0.896                   & 31.43, 0.934                    \\
BIRNAT                            & 29.50, 0.900                   & 27.92, 0.922                   & 36.86, 0.970                   & 41.22, 0.989                   & 27.80, 0.903                   & 28.34, 0.897                   & 31.94, 0.930                    \\
RevSCI                            & 28.47, 0.854                   & 27.97, 0.921                   & 37.04, 0.970                   & 41.43, 0.989                   & 27.64, 0.896                   & 28.69, 0.903                   & 31.87, 0.922                    \\
STFormer                          & 30.55, 0.946                   & 29.42, 0.952                   & 38.76, 0.983                   & 42.04, 0.992                   & 28.72, 0.950                   & 29.79, 0.938                   & 33.21, 0.960                    \\
EfficientSCI                      & 30.83, 0.949                   & 29.60, 0.955                   & 38.50, 0.983                   & 42.34, 0.992                   & 29.05, 0.955                   & 29.53, 0.936                   & 33.31, 0.962                    \\
RegDif(Ours)                      & 33.47, 0.971                   & 31.32, 0.967                   & 41.23, \underline{0.988}           & 45.28, 0.995                   & 29.58, 0.963                   & 30.46, 0.949                   & 35.22, 0.972                    \\ 
\hline
DoP-EfficientSCI                  & \underline{35.51}, \underline{0.974}   & \underline{32.03}, \underline{0.973}   & \underline{41.38}, 0.984           & \underline{45.72}, \underline{0.995}   & \underline{30.08}, \underline{0.978}   & \underline{31.76}, \underline{0.951}   & \underline{36.08}, \underline{0.976}    \\
DoP-RegDif(Ours)                  & \textbf{37.14}, \textbf{0.984} & \textbf{33.72}, \textbf{0.979} & \textbf{44.45}, \textbf{0.989} & \textbf{47.05}, \textbf{0.996} & \textbf{31.69}, \textbf{0.979} & \textbf{34.41}, \textbf{0.975} & \textbf{38.08}, \textbf{0.984}  \\
\hline
\end{tabular}
\end{adjustbox}
\caption{PSNR (left) and SSIM (right) on six grayscale datasets under single and dual optical path (DoP) with one-hot mask.}
\label{tab:grayscale}
\end{table*}

\begin{table*}
\centering
\begin{adjustbox}{width=0.955\textwidth}
\begin{tabular}{l|c|c|c|c|c|c|c} 
\hline
\multicolumn{1}{c|}{Color Datasets} & ShakeNDry                      & Traffic                        & Jockey                         & Beauty                         & Runner                         & Bosphorus                      & Average                         \\ 
\hline\hline
GAP-TV                             & 22.58, 0.421                   & 19.94, 0.546                   & 24.73, 0.521                   & 19.75, 0.327                   & 28.00, 0.806                   & 24.30, 0.501                   & 23.22, 0.520                    \\
PnP-FFDNet                         & 33.19, 0.943                   & 24.18, 0.856                   & 34.00, 0.946                   & 35.33, 0.971                   & 34.79, 0.951                   & 32.93, 0.954                   & 32.41, 0.937                    \\
PnP-FastDVDnet                     & 33.98, 0.946                   & 26.55, 0.912                   & 35.13, 0.952                   & 36.05, 0.973                   & 37.22, 0.973                   & 37.56, \underline{0.978}           & 34.41, 0.956                    \\
EfficientSCI                       & 34.71, 0.934                   & 28.79, 0.913                   & 37.89, 0.946                   & 36.84, 0.966                   & 41.75, 0.982                   & 40.30, 0.967                   & 36.71, 0.951                    \\
RegDif(Ours)                       & 35.71, \underline{0.947}           & 30.34, \underline{0.937}           & 38.51, 0.954                   & 37.22, \underline{0.978}           & 42.09, 0.985                   & 40.48, 0.972                   & 37.39, \underline{0.962}            \\ 
\hline
DoP-EfficientSCI                   & \underline{35.73}, 0.944           & \underline{30.47}, 0.933           & \underline{38.67}, \underline{0.954}   & \underline{37.43}, 0.975           & \underline{42.20}, \underline{0.988}   & \underline{40.81}, 0.971           & \underline{37.55}, 0.961            \\
DoP-RegDif(Ours)                   & \textbf{36.17}, \textbf{0.954} & \textbf{30.86}, \textbf{0.949} & \textbf{39.17}, \textbf{0.969} & \textbf{37.73}, \textbf{0.986} & \textbf{43.74}, \textbf{0.989} & \textbf{41.74}, \textbf{0.979} & \textbf{38.24}, \textbf{0.971}  \\
\hline
\end{tabular}
\end{adjustbox}
\caption{PSNR (left) and SSIM (right) on six color datasets under single and dual optical path (DoP) with one-hot mask.}
\label{tab:color}
\end{table*}

On simulated grayscale datasets, we compare RegDif with model-based methods (GAP-TV \cite{yuan2016generalized}, PnP-FFDNet \cite{yuan2020plug}, PnP-FastDVDnet \cite{yuan2021plug}) and deep learning-based methods (BIRNAT \cite{cheng2020birnat}, RevSCI \cite{cheng2021memory}, STFormer \cite{wang2022spatial}, EfficientSCI \cite{wang2023efficientsci}) in the single-path configuration. For dual path, we compare DoP-RegDif (RegDif enhanced by compensatory-path measurement) with DoP-EfficientSCI (EfficientSCI enhanced through the same fusion method as DoP-RegDif). In terms of quantitative comparisons shown in Table~\ref{tab:grayscale}, our RegDif outperforms previous model-based methods by more than \textbf{3.5} dB within the single-path setting. Compared to EfficientSCI, RegDif achieves an improvement of \textbf{1.9} dB while maintaining comparable parameters (EfficientSCI: 8 regression blocks; RegDif: 1 regression + 7 diffusion blocks; same 256 hidden dimension with similar architecture). In the dual-path setting, DoP-RegDif yields \textbf{2.8} dB over RegDif by incorporating additional measurement and outperforms DoP-EfficientSCI by \textbf{2.0} dB. As shown in Fig.~\ref{fig:sim_result}, our method retrieves more details and textures in grayscale datasets.

On simulated color datasets, reconstruction is more challenging due to spatial masking interacting with the Bayer filter. Beyond changing output channels from 1 to 3, RegDif requires additional masked-region processing during coarse-prediction updates (details are provided in Appendix D). We compare RegDif with GAP-TV, PnP-FFDNet, PnP-FastDvDnet, and EfficientSCI in the single-path configuration and compare DoP-RegDif with DoP-EfficientSCI for dual path. Our RegDif outperforms EfficientSCI by \textbf{0.68} dB under the single-path setting, while DoP-RegDif outperforms DoP-EfficientSCI by \textbf{0.69} dB under the dual-path setting. As shown in Fig.~\ref{fig:sim_result}, our method is better at restoring accurate colors and fine structures in color datasets.

\subsection{Results on Real Captured Videos}
In real-world scenarios, the presence of noise often leads to discrepancies compared to simulations. RegDif, which introduces Gaussian noise in the SDE, provides advantages in reconstructing real scenes. Furthermore, the limited bit depth of the camera causes the restricted dynamic range; in this context, our one-hot mask demonstrates a clear advantage over the random binary mask. We employ a dual optical path with the one-hot mask to capture a dynamic scene featuring a rotating disc displaying four digits, each situated in a region of a different color. After obtaining two complementary measurements, we reconstruct the dynamic scene with DoP-RegDif, and the results are illustrated in Fig.~\ref{fig:real_result}.

\subsection{Ablation Study}
To offer insights into the proposed method, we analyze the impact of loss functions in our framework by recording two metrics during training: the $\ell_2$-norm distance between ground truth and One-step Denoiser output (final output), as well as the distance between ground truth and Regressor Initializer output (coarse output) under three conditions: ($i$) all three losses are utilized, ($ii$) the Alignment Loss is omitted, and ($iii$) the Regression Loss is removed. As shown in Fig.~\ref{fig:ablation_loss}, the final output under condition ($i$) performs the best. When the Alignment Loss is absent (which aligns the coarse output within the diffusion process), the lack of this constraint enhances the coarse output but degrades the final output compared to condition ($i$). Furthermore, when ablating Regression Loss (which constrains the coarse result), the Alignment Loss may cause deterioration of the coarse output by aligning it within reverse process, leading to a decline in the final output relative to condition ($i$). Additional details, including ablation studies of model architecture, inference times, and parameter sizes are provided in Appendix E.

\section{Conclusion}
% Leveraging the one-hot mask's ability to effectively distinguish information between different frames in the modulated measurement, we are the first to transform the regression problem in video SCI into a diffusion-based inpainting problem by proposing an SDE of the forward process that aligns with the hardware compression process. To tackle the challenges of slow reconstruction speed and poor quality associated with the pure diffusion method in video SCI, we propose a hybrid reconstruction approach named RegDif, which employs regression to predict coarse results and subsequently applies one-step denoising. Furthermore, we design a dual path at the hardware level to mitigate the degradation caused by the one-hot mask by incorporating an additional measurement. Extensive experiments on grayscale, color, and real data demonstrate the superiority of our method under the one-hot modulation configuration.

Leveraging the one-hot mask's ability to effectively distinguish information between frames in modulated measurements, we are the first to transform the regression problem in video SCI into a diffusion-based inpainting problem by proposing an SDE of the forward process that aligns with the hardware compression process. To address slow reconstruction speed and poor quality of the pure diffusion in video SCI, we propose a hybrid framework named RegDif, using one-step regression to predict coarse results followed by one-step denoising. Furthermore, we design a hardware-level dual path incorporating an additional measurement to mitigate degradation caused by one-hot masks. Extensive experiments on grayscale, color, and real data demonstrate our method's superiority under one-hot modulation.

\section{Acknowledgments}
\thanks{This work was supported by the National Key R\&D Program of China under Grant 2024YFF0505603, National Natural Science Foundation of China under Grant 62271414, Zhejiang Provincial Science Fund for Distinguished Young Scholar Project under Grant LR23F010001, Zhejiang ``Pioneer" and ``Leading Goose" R\&D Program under Grant 2024SDXHDX0006 and 2024C03182, the Key Project of Westlake Institute for Optoelectronics under Grant 2023GD007, and Ningbo Science and Technology Bureau, ``Science and Technology Yongjiang 2035" Key Technology Breakthrough Program under Grant 2024Z126.}
\bibliography{aaai2026}

\end{document}